# Lightweight Multimodal Adaptation of Vision–Language Models for Species Recognition and Habitat-Context Interpretation in Drone Thermal Imagery

Hao Chen[a], Fang Qiu[a*], Fangchao Dong[a], Defei Yang[a], Eve Bohnett[b], Li An[c,d]

a. Geospatial Information Science, The University of Texas at Dallas, Richardson, TX 75080, USA

b. Department of Landscape Architecture, University of Florida, Gainesville, FL 32611, USA

c. College of Forestry, Wildlife and Environment, Auburn University, Auburn, AL 36849, USA

d. International Center for Climate and Global Change Research, College of Forestry, Wildlife and Environment, Auburn University, Auburn, AL 36849, USA

Corresponding author: Fang Qiu, Geospatial Information Sciences, University of Texas at Dallas, 800 West Campbell Road, Richardson, TX, 75080, United States, ffqiu@utdallas.edu

**Abstract**: This study proposes a lightweight multimodal adaptation framework to bridge the representation gap between RGB-pretrained VLMs and thermal infrared imagery, and demonstrates its practical utility using a real drone-collected dataset. A thermal dataset was developed from drone-collected imagery and was used to fine-tune VLMs through multimodal projector alignment, enabling the transfer of information from RGB-based visual representations to thermal radiometric inputs. Three representative models, including InternVL3-8B-Instruct, Qwen2.5-VL-7B-Instruct, and Qwen3-VL-8B-Instruct, were benchmarked under both closed-set and open-set prompting conditions for species recognition and instance enumeration. Among the tested models, Qwen3-VL-8B-Instruct with open-set prompting achieved the best overall performance, with F1 scores of 0.935 for deer, 0.915 for rhino, and 0.968 for elephant, and within-1 enumeration accuracies of 0.779, 0.982, and 1.000, respectively. In addition, combining thermal imagery with simultaneously collected RGB imagery enabled the model to generate habitat-context information, including land-cover characteristics, key landscape features, and visible human disturbance. Overall, the findings demonstrate that lightweight projector-based adaptation provides an effective and practical route for transferring RGB-pretrained VLMs to thermal drone imagery, expanding their utility from object-level recognition to habitat-context interpretation in ecological monitoring.

# 1. Introduction

Vision–language models (VLMs) have emerged as a powerful paradigm for multimodal understanding by linking visual representations with natural language generation and reasoning. However, most existing VLMs are predominantly pretrained on web-scale RGB image–text pairs, and their capabilities have been developed and evaluated primarily in RGB-centered settings (Radford et al., 2021; Liu et al., 2023; Fu et al., 2023). Thermal infrared imagery serves as a critical complementary sensing modality to RGB data by capturing emitted radiance rather than reflected light, and has been widely applied in ecological monitoring (Still et al., 2019), search and rescue (Yeom, 2024), precision agriculture (Khanal et al., 2017), and medical thermography (Lahiri et al., 2012).

Nevertheless, as illustrated in Fig. 1, even state-of-the-art VLMs exhibit significant challenges in species identification and instance counting when applied to thermal imagery. Thermal imagery encodes emitted radiance as single-channel intensity, lacking color, fine texture, and detailed morphological cues. This results in reduced inter-class separability and limited structural definition, making species discrimination and instance counting significantly more challenging. In addition, limited spatial resolution and the absence of reliable scale cues introduce ambiguity in instance counting, particularly when multiple individuals are spatially clustered or partially occluded. These limitations reflect a fundamental cross-modal discrepancy: representations learned from RGB imagery are not directly transferable to thermal infrared data, requiring explicit alignment of radiometric and semantic features beyond conventional domain adaptation. Extending VLMs from RGB to thermal infrared imagery is therefore important not only for domain-specific applications but also for advancing cross-modal generalization in multimodal artificial intelligence.

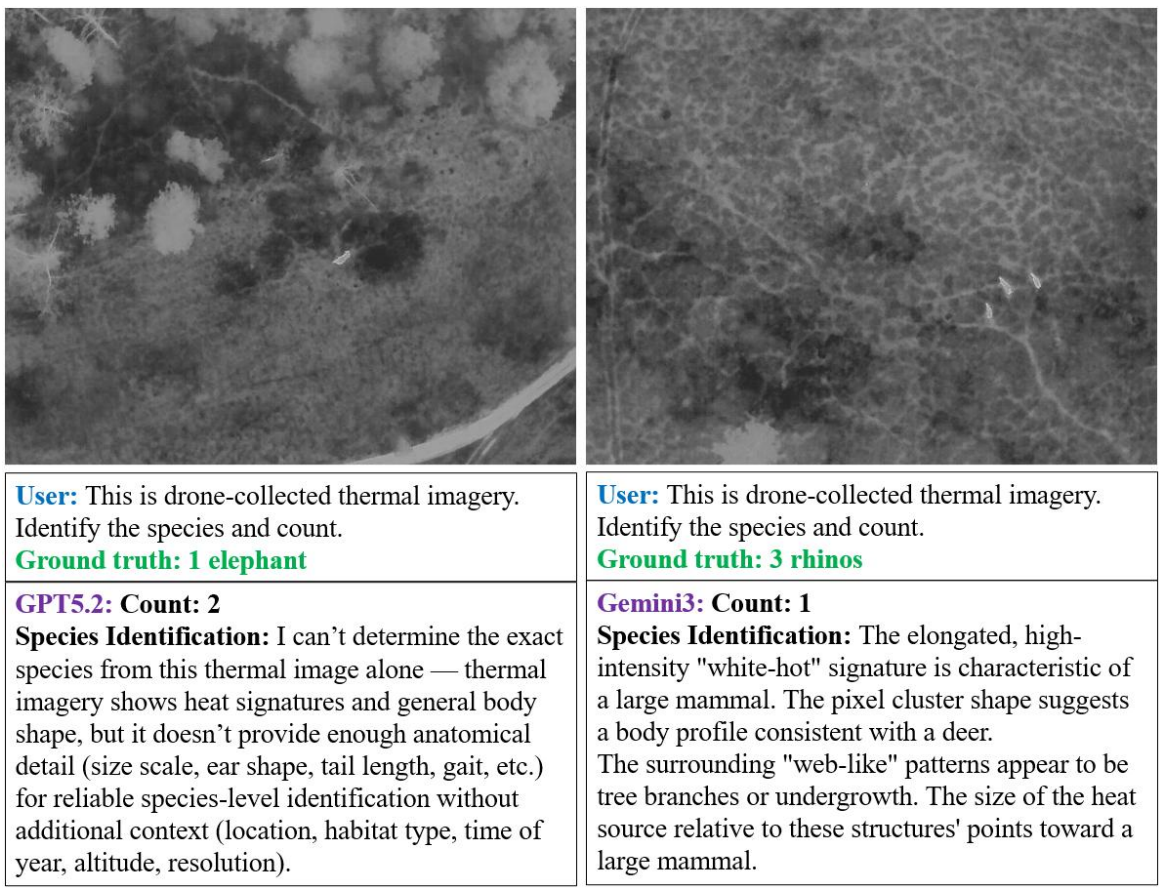

**Figure 1**. Representative failures of foundational VLMs in species recognition and instance enumeration from drone-collected thermal imagery.

Despite the growing importance of thermal sensing, relatively few studies have systematically investigated how to bridge the representation gap between RGB-pretrained VLMs and thermal infrared imagery. Existing work has mainly focused either on diagnosing RGB–thermal understanding gaps through benchmark-based evaluation or on building infrared-specific datasets and models through data-intensive training. For example, RGB-Th-Bench (Moshtaghi et al., 2025) establishes a benchmark for evaluating VLMs in RGB–thermal settings and reveals significant limitations in thermal understanding; however, it is primarily diagnostic and does not address how to adapt RGB-pretrained VLMs to thermal imagery. IRGPT (Cao et al., 2025) introduces a large-scale real-world infrared image–text dataset and a bi-cross-modal curriculum transfer learning framework for infrared vision–language learning; however, it relies on data-intensive multi-stage training to develop an infrared-specific model rather than enabling lightweight adaptation of general-purpose VLMs. As a result, parameter-efficient multimodal alignment for transferring RGB-pretrained VLMs to thermal imagery remains largely underexplored.

In ecological monitoring in particular, drone-collected thermal imagery has demonstrated substantial value for wildlife observation, especially by improving the detection of animals under low-light conditions and beneath partial tree cover (Anderson et al., 2025; Bohnett et al., 2025; dos Santos et al., 2025; Meade et al., 2025). Compared to conventional thermal imaging, drone-collected thermal imagery presents additional challenges, including top-down observation geometry, small object representations, limited spatial resolution, and significant radiometric variability due to environmental conditions. These factors further reduce discriminative visual cues, increasing the difficulty of reliable species recognition and instance counting.

To address these challenges, existing studies have primarily relied on deep learning–based object detectors, such as YOLO and Faster R-CNN, which have achieved considerable improvements in detection accuracy and computational efficiency (Lyu et al., 2024; Vavekanand et al., 2026). However, these approaches are inherently limited to object-level outputs, such as bounding boxes and class probabilities, and do not capture higher-level environmental semantics or contextual information that are critical for ecological interpretation.

To address these limitations, this study proposes a lightweight, parameter-efficient multimodal adaptation framework to bridge the representation gap between RGB-pretrained VLMs and thermal infrared imagery. Specifically, a drone-collected thermal dataset is constructed to adapt VLMs via multimodal projector alignment. Only the visual projection module is trained, while the backbone networks remain frozen. This approach enables effective transfer from RGB-based visual representations to thermal radiometric inputs with minimal computational cost. Three representative models— InternVL3-8B-Instruct (Zhu et al., 2025), Qwen2.5-VL-7B-Instruct (Bai et al., 2025a), and Qwen3-VL-8B-Instruct (Bai et al., 2025b)— are systematically evaluated under both closed-set and open-set prompting conditions for species recognition and instance enumeration. Furthermore, by integrating thermal imagery with simultaneously acquired RGB data, the framework is extended to support habitat-context interpretation, including land-cover characterization, identification of key landscape features, and detection of human disturbance, leveraging the semantic reasoning capabilities of VLMs.

This study makes three key contributions: First, we introduce a parameter-efficient multimodal alignment strategy that enables RGB-pretrained VLMs to operate effectively in the thermal radiometric domain for species recognition and counting. Second, we demonstrate that fine-tuned VLMs adapted to thermal imagery achieve competitive or superior performance on semantic-level tasks, including species identification and instance enumeration, compared to conventional deep learning detector-based pipelines. Third, we establish a unified VLM-based framework that extends beyond object detection to structured ecological information extraction, supporting scalable and human-interactive environmental monitoring.

## 2. Related Work

### 2.1 Drone Thermal Imagery for Wildlife Monitoring

Drone-based thermal imaging has become an important tool for wildlife monitoring by enabling rapid surveys over large areas and improving the detection of animals in low-light conditions and in visually obstructed habitats. Recent studies show that thermal drones are now used across a broad range of ecological applications, with population surveys remaining the dominant use case (Rahman et al., 2025; Norris et al., 2026). The application scope of thermal drone imagery extends beyond large mammals to include smaller or more difficult-to-observe taxa, such as birds and bats (Pinel-Ramos et al., 2025; McCarthy et al., 2021; Santangeli et al., 2020).

Because thermal imaging remains effective at night, it has enabled research beyond simple presence–absence detection. In addition to ecological surveys, thermal drones have been applied in nocturnal anti-poaching surveillance, where thermal infrared imagery demonstrates a higher human detection probability than RGB imagery, particularly under favorable thermal contrast conditions (Hambrecht et al., 2019). More broadly, drone-based thermal imaging is increasingly being used to support animal behavior research (Pedrazzi et al., 2025). For example, recent studies have investigated sleeping-site selection in arboreal primates using thermal drones (Gazagne et al., 2025). Earlier work demonstrated that thermal imaging can support the study of sleeping behavior in critically endangered Hainan gibbons in natural habitats (Zhang et al., 2020), as well as monitor night-time behavior of European brown hares (*Lepus europaeus*) (Povlsen et al., 2022).

Beyond detection and behavioral observation, thermal sensing also provides additional ecological information due to its ability to capture emitted radiance and thermal contrast patterns. This capability enables applications in wildlife health and management. For example, drone-based thermal imaging has been used to detect wild boar carcasses for disease management, supporting efforts to locate reservoirs of infection in the landscape (Rietz et al., 2023). In addition, drone-based infrared thermography has been validated as a non-invasive approach for estimating dolphin vital signs (White et al., 2025), further demonstrating its potential for physiological monitoring.

To support efficient analysis of large-scale drone thermal imagery, deep learning-based detectors have been applied for automated wildlife detection and identification (Xu et al., 2024). Recent research mainly targets better detection through enhanced small-target recognition, real-time aerial-video processing, and combining visible and thermal imagery for more robust classification. Representative studies illustrate these strategies. For instance, Lyu et al. (2024) improved small-object detection in thermal imagery by enhancing Faster R-CNN with feature pyramid networks and customized anchors, while Povlsen et al. (2023) employed YOLOv5 for real-time detection of hares and roe deer in aerial thermal video. He et al. (2024) proposed ALSS-YOLO, a YOLOv8-based lightweight network that integrates channel splitting and shuffling mechanisms to enhance feature representation and improve detection efficiency for small thermal targets in UAV imagery. Additionally, Krishnan et al. (2023) evaluated eight visible–thermal image-fusion methods with YOLOv5 and YOLOv7, showing that multimodal fusion can improve automated animal detection and classification in drone surveys, particularly for cryptic species such as white-tailed deer.

Detector-based pipelines remain limited in their ability to support higher-level ecological interpretation. Most current methods are optimized for object-level outputs such as presence, class labels, coordinates, or bounding boxes, which are valuable for census-oriented tasks but provide limited semantic information about habitat conditions, landscape context, or signs of human disturbance. In practice, these outputs often require substantial post-processing or expert interpretation to support richer ecological analysis, and their performance is further challenged by small targets, occlusion, scale variation, and cluttered thermal backgrounds. These limitations highlight the need for vision–language-based multimodal frameworks that can move beyond object detection to support structured ecological interpretation, including species identification, instance counting, and habitat-context reasoning.

## 2.2 Vision-language Models

VLMs enable joint visual understanding and language reasoning by learning aligned multimodal representations. Owing to their strong multimodal comprehension capabilities, VLMs have developed rapidly in recent years. Unlike conventional deep learning-based detectors that rely on predefined categories and produce limited object-level outputs, VLMs operate in a shared vision–language semantic space, enabling open-vocabulary recognition and more flexible interpretation of visual content. This allows models to move beyond object localization toward richer semantic understanding, including structured descriptions and context-aware reasoning.

A representative line of VLM development is contrastive vision–language pretraining, as exemplified by Contrastive Language–Image Pre-training (CLIP) (Radford et al., 2021), which aligns images and text in a shared semantic space and supports zero-shot recognition and open-vocabulary reasoning. Building on this foundation, RemoteCLIP adapts CLIP to the remote-sensing domain by learning image–text alignment from remote-sensing data and achieves average zero-shot classification improvements of up to 6.39% over the original CLIP baseline, demonstrating the effectiveness of contrastive VLMs for drone imagery (Liu et al., 2024). In addition to recognition tasks, CLIP-based models have also been extended to more complex remote-sensing applications. For example, ChangeCLIP introduces multimodal vision–language representation learning for remote sensing change detection, enabling the identification of semantic changes across multi-temporal imagery (Dong et al., 2024). Furthermore, recent studies have further extended CLIP-like models to open-vocabulary semantic segmentation in drone imagery, indicating that these models can support more detailed scene understanding in drone-based applications (Chen et al., 2025). Zhou et al. (2025) further highlight open-vocabulary analysis as a promising direction for drone imagery, particularly for applications requiring more flexible and intelligent aerial perception, such as emergency rescue, wildlife protection, and smart-city management.

Unlike CLIP-style contrastive models, which focus on dual-encoder alignment for recognition-oriented tasks, Bootstrapping Language–Image Pre-training (BLIP) (Li et al., 2023) is an encoder–decoder vision–language framework designed for both understanding and language generation. This design makes BLIP-style models particularly relevant to remote-sensing image captioning, where the goal is not only to recognize scene elements but also to generate coherent semantic descriptions. Building on BLIP-style vision–language pretraining, Yang et al. (2024) proposed a two-stage alignment framework to improve image–text interaction for remote-sensing captioning, while Zhao et al. (2025) explored parameter-efficient transfer learning to adapt BLIP-like models to this domain with reduced computational cost. Beyond captioning, BLIP-related vision–language methods have also expanded to other cross-modal remote-sensing tasks. RSGPT (Hu et al., 2025) is built on the InstructBLIP framework and introduces domain-specific captioning and evaluation datasets for remote-sensing image captioning and visual question answering, while Knowledge-Aware Text–Image Retrieval (KTIR) (Mi et al., 2024) extends BLIP-related cross-modal retrieval by incorporating external knowledge graphs to enrich text queries and improve image–text matching. More broadly, Remote Sensing ChatGPT incorporates BLIP-like multimodal perception modules within large language model pipelines to support interactive, language-guided remote-sensing analysis (Guo et al., 2024).

Despite the rapid development of both drone thermal detection methods and vision–language models for remote sensing, several limitations remain. Current approaches remain largely centered on detection and counting, with limited capacity for higher-level ecological interpretation. Meanwhile, CLIP- and BLIP-based remote-sensing VLMs have largely been developed on RGB imagery, leaving the representation gap between RGB-pretrained models and thermal infrared data insufficiently addressed. Furthermore, many existing multimodal approaches depend on full-model retraining, large-scale domain-specific data, or task-specific architectures, reducing their flexibility and computational efficiency. These gaps highlight the need for a lightweight multimodal adaptation strategy that can effectively transfer RGB-pretrained VLMs to thermal drone imagery for species recognition, instance enumeration, and habitat-context understanding.

## 3. Methodology

In this study, thermal and RGB imagery were collected simultaneously using drone platforms. Thermal imagery served as the primary modality for vision–language model training and evaluation, whereas RGB imagery was used as a visual reference for annotation and as complementary contextual information for habitat-context interpretation. A thermal image dataset was constructed for multimodal projector alignment, enabling RGB-pretrained VLMs to be adapted to thermal infrared imagery. Comparative benchmarking was conducted using InternVL3-8B-Instruct, Qwen2.5-VL-7B-Instruct, and Qwen3-VL-8B-Instruct under both closed-set and open-set prompting settings. The best-performing model was then selected for detailed analysis of species recognition and habitat-context interpretation. The overall workflow of the proposed framework is shown in Fig. 2.

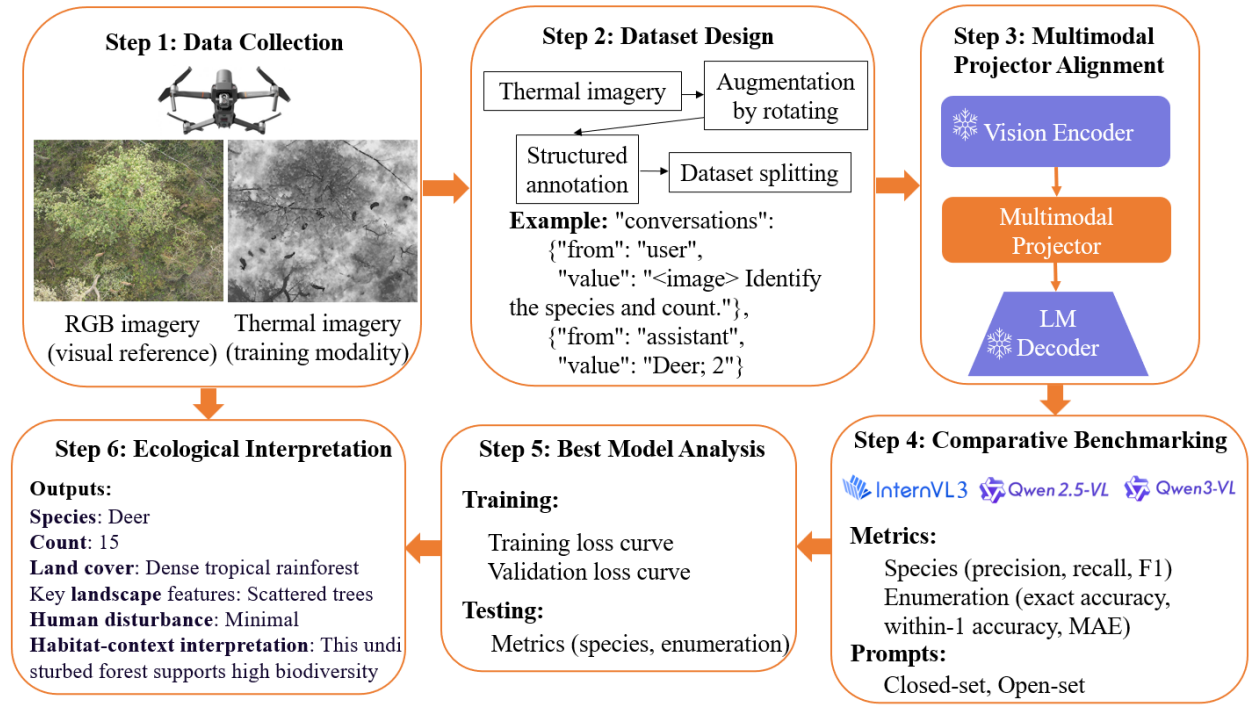

**Figure 2**. Flowchart of the proposed method.

### 3.1 Data Collection

The data were collected in Chitwan National Park (CNP) and its surrounding regions (Fig. 3). As the first national park in Nepal, CNP was designated a World Heritage Site in 1984, due to its internationally significant ecosystems (Mishra, 1982). The park covers a variety of ecosystems and landscape features, including extensive grasslands, subtropical deciduous forests, riverine systems, wetlands, and adjacent hill landscapes, all of which provide important habitat for wildlife. CNP supports 68 mammal species and contains the largest population of one-horned rhinoceroses (rhinos) in Nepal (UNESCO, 2019). The park also sustains a diverse assemblage of deer species, which are ecologically important herbivores widely distributed across both grassland and forest habitats.

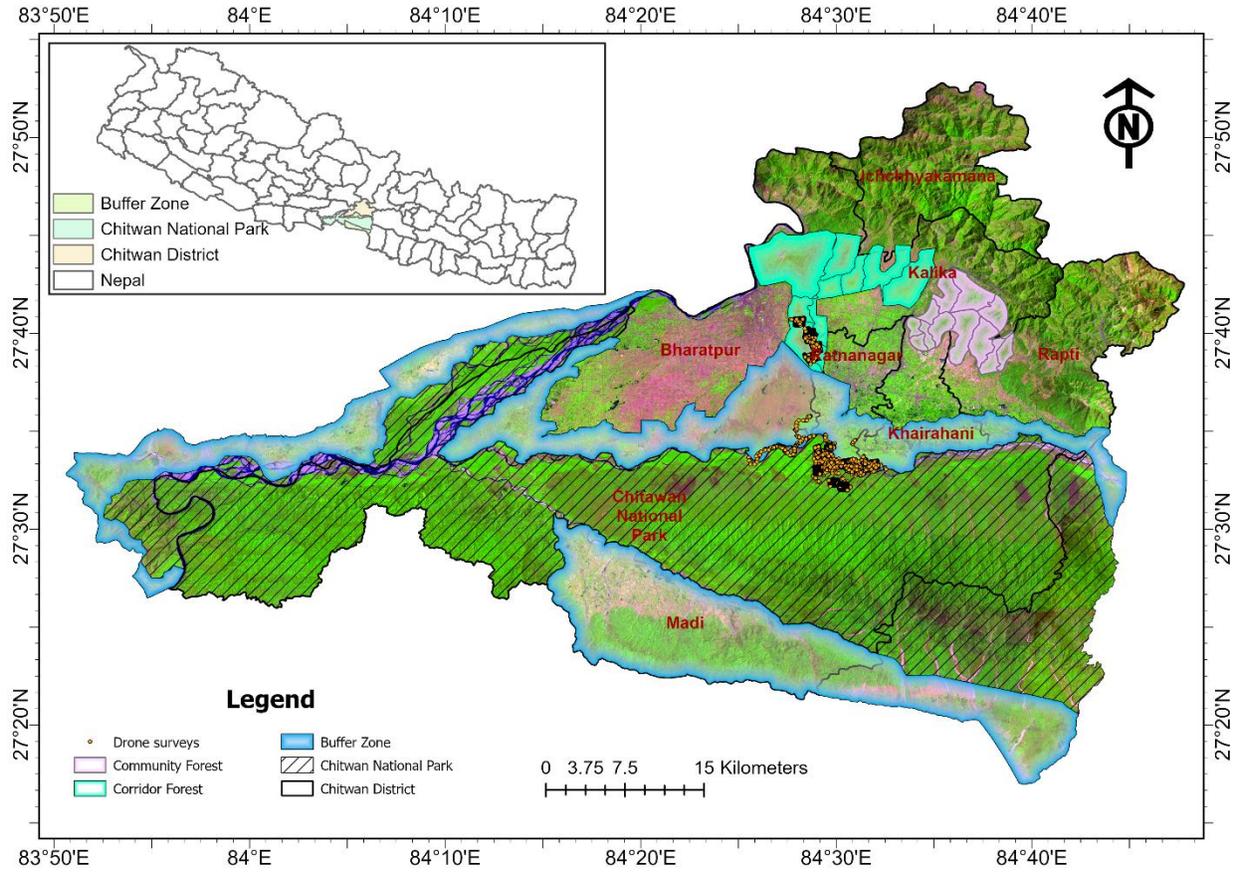

**Figure 3**. Drone survey locations in the study area (False color image of Landsat 8 collected on April 5, 2022. Red: band 6, 1560-1660 nm; Green: band 5, 845-885 nm; Blue: band 4, 630-680 nm).

Drone imagery was acquired across the study area between February and July 2022 using a DJI Mavic 2 Enterprise platform equipped with both thermal and visible RGB sensors. Both sensors were stabilized by a three-axis mechanical gimbal, which improved image stability and consistency during flight. The system simultaneously recorded thermal infrared and RGB imagery, with the thermal images captured at a resolution of 640 × 512 pixels and RGB images at 8000 × 6000 pixels. Thermal imagery served as the primary data source for species recognition, instance enumeration, and model evaluation, whereas RGB imagery was used for visual verification, annotation support, and habitat-context interpretation.

### 3.2 Dataset Construction

To support the multimodal projector alignment for VLMs, drone thermal imagery containing elephants, rhinos, and deer were selected to construct the dataset. The dataset was organized in a conversational annotation format following the ShareGPT structure. Each image was paired with a user–assistant supervision example, in which the user prompt instructed the model <*to identify the species and count*>, and the assistant returned a concise structured response (e.g., *Elephant; 2*). Representative of these user-assistant annotation pairs are illustrated in Fig. 4, illustrating the conversational supervision used to align the multimodal projector with species semantics and instance counts.

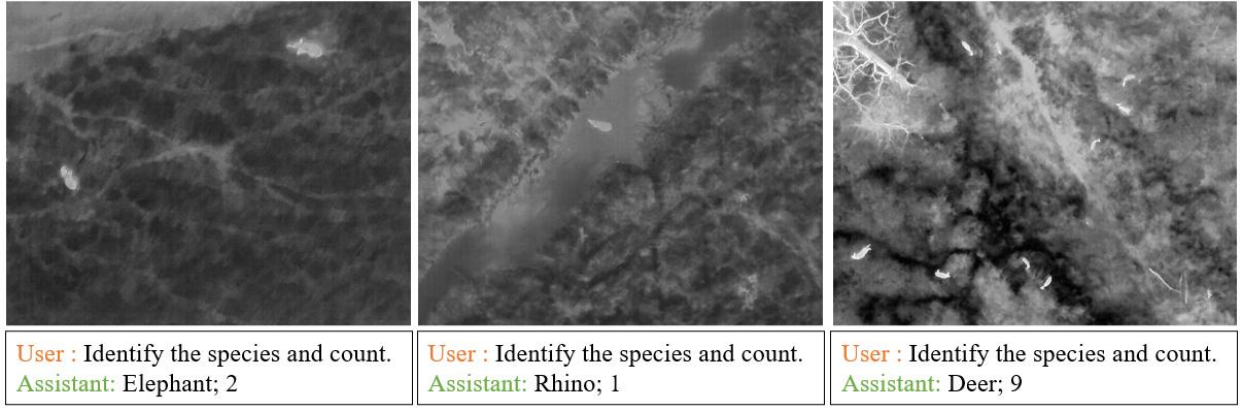

**Figure 4**. Structured annotation in the drone thermal dataset.

To address class imbalance, underrepresented classes, including rhinos and elephants, were augmented through random rotation. Species recognition and instance enumeration were annotated by ecology specialists, and all annotations were further cross-validated by experts to ensure accuracy and consistency. The resulting dataset contained 2,266 images per species. Finally, the dataset was split into training, validation, and test subsets at proportions of 80%, 10%, and 10%, respectively.

**3.3 Multimodal Projector Alignment**

Vision–language models typically contain billions of parameters, making full fine-tuning computationally expensive and often impractical. Moreover, most existing foundation VLMs are pretrained predominantly on web-scale RGB image–text pairs rather than thermal infrared data. As a result, although the pretrained vision encoder and language model capture rich visual and semantic knowledge, the learned cross-modal representations are not naturally aligned with the radiometric characteristics of thermal imagery.

To address the modality mismatch, this study adopts multimodal projector alignment as a lightweight adaptation strategy for thermal imagery. Instead of updating the entire VLM, only the projector layer linking the vision encoder and the language model is optimized, while both the vision tower and the language backbone remain frozen during training. This design enables thermal visual features to be mapped into the multimodal representation space of RGB-pretrained VLMs in a parameter-efficient manner. The cross-modal alignment was trained on a class-balanced drone thermal dataset using a token-level autoregressive language-modeling objective, in which the model was optimized to generate a target response sequence containing the annotated species and instance count (Liu et al., 2023). The training objective is defined as

$$\mathcal{L}_{proj} = -\sum_{t=1}^{T} logP(w_t | w_{<t}, z_{th}) \quad (1)$$

where $w_t$ denotes the *t*-th token in the target response sequence, $w_{<t}$ represents all preceding tokens, $z_{th}$ denotes the projected thermal visual representation, and $P(w_t|w_{<t}, z_{th})$ is the conditional probability of the *t*-th token predicted by the language model given the previous tokens and the projected thermal representation. Under this objective, the projector learns to align thermal visual features with the semantic space of the pretrained VLM, enabling the model to generate outputs for species recognition and instance enumeration.

In this study, three state-of-the-art VLMs were adapted using multimodal projector alignment: InternVL3-8B-Instruct (Zhu et al., 2025), Qwen2.5-VL-7B-Instruct (Bai et al., 2025), and Qwen3-VL-8B-Instruct (Bai et al., 2025). As shown in Table 1, only a small fraction of model parameters was updated in each case, while the pretrained vision and language backbones remained frozen. This configuration highlights the parameter-efficient design of the proposed adaptation framework. Among the three models, InternVL3-8B-Instruct had the smallest proportion of trainable parameters, whereas Qwen3-VL-8B-Instruct had the largest.

**Table 1** Fine-tuning parameter configuration.

| Foundation model | Trained parameters | Trained (%) | Trained tensors | Frozen tensors | Trainable modules |
|---|---|---|---|---|---|
| Qwen2.5-VL-7B-Instruct | 44,574,464 | 0.537% | 5 | 724 | Merger/MLP |
| Qwen3-VL-8B-Instruct | 160,496,896 | 1.831% | 24 | 726 | Merger/Deepstack |
| InternVL3-8B-Instruct | 27,540,480 | 0.347% | 6 | 679 | MLP |

### 3.4 Prompts and Metrics

To evaluate the influence of prompting strategy on model inference, two prompting settings were designed: closed-set prompting and open-set prompting. In the closed-set setting, the model was explicitly restricted to three candidate species (deer, rhino, and elephant). In the open-set setting, no species constraints were provided, allowing the model to generate species names freely while preserving the same structured output format.

The closed-set prompt was defined as:
*Identify the species and count. Return ONLY in the format: Species; Count (example: Deer; 1). Allowed species: deer, rhino, elephant.*

The open-set prompt was defined as:
*Identify the species and count. Return ONLY in the format: Species; Count (example: Deer; 1).*

For species recognition, model performance was assessed using precision, recall, and F1-score for each species. For instance, enumeration, performance was evaluated using exact count accuracy, within-1 accuracy, and mean absolute error (MAE). Exact count accuracy reflects the proportion of predictions in which the count exactly matched the reference value. Within-1 accuracy reflects the proportion of predictions with an absolute counting error less than or equal to one. MAE measures the average absolute deviation between predicted and reference counts. Together, these metrics provide a comprehensive evaluation of species recognition and instance enumeration performance (Norouzzadeh et al., 2018).

### 4. Experiments

Three state-of-the-art VLMs were adapted using multimodal projector alignment: InternVL3-8B-Instruct (Zhu et al., 2025), Qwen2.5-VL-7B-Instruct (Bai et al., 2025), and Qwen3-VL-8B-Instruct (Bai et al., 2025). All adapted models were initialized from their pretrained foundation weights, and the corresponding foundation models were used as baselines. Training was conducted using the AdamW optimizer with BFloat16 mixed-precision to improve computational efficiency and reduce memory consumption. The training configuration used an initial learning rate of $1 \times 10^{-4}$, a warmup ratio of 0.03, a maximum of 1,000 training steps, and a cosine learning-rate scheduler, with optimization implemented using *adamw_torch*. To ensure a fair and consistent experimental setting, all training and evaluation were performed on a single NVIDIA GeForce RTX 3090 GPU with 24 GB of VRAM, further highlighting the parameter-efficient nature of the proposed strategy.

#### 4.1 Comparative Benchmarking

#### 4.1.1 Species Recognition

Table 2 presents species recognition performance for both the foundation models and the projector-aligned models under closed-set and open-set prompting. Overall, multimodal projector alignment markedly improved recognition performance across all three VLMs. The untuned foundation models showed limited capability on thermal imagery, particularly for rhino and elephant, whereas the tuned models achieved substantial gains in precision, recall, and F1-score across all species.

**Table 2** Comparison of species recognition performance across models.

| Model | Prompt | Precision | | | Recall | | | F1 | | |
| --- | --- | --- | --- | --- | --- | --- | --- | --- | --- | --- |
| | | Deer | Rhino | Elephant | Deer | Rhino | Elephant | Deer | Rhino | Elephant |
| Qwen2.5-VL-7B-Instruct | Closed-Set | 0.329 | 0.500 | 0.750 | 0.947 | 0.004 | 0.013 | 0.489 | 0.009 | 0.260 |
| | Open-Set | 0.280 | 1.000 | 0.500 | 0.062 | 0.004 | 0.004 | 0.101 | 0.009 | 0.009 |
| Qwen3-VL-8B-Instruct | Closed-Set | 0.336 | 0.529 | 0.667 | 0.903 | 0.040 | 0.053 | 0.490 | 0.074 | 0.098 |
| | Open-Set | 0.431 | 0 | 0.750 | 0.097 | 0 | 0.013 | 0.159 | 0 | 0.026 |
| InternVL3-8B-Instruct | Closed-Set | 0.333 | 0 | 0 | 1.000 | 0 | 0 | 0.500 | 0 | 0 |
| | Open-Set | 0.329 | 0 | 0 | 0.252 | 0 | 0 | 0.286 | 0 | 0 |
| Qwen2.5-VL-7B-Instruct-Tuned | Closed-Set | 0.823 | 0.744 | 0.784 | 0.907 | 0.681 | 0.770 | 0.863 | 0.711 | 0.777 |
| | Open-Set | 0.910 | 0.893 | 0.805 | 0.889 | 0.668 | 0.929 | 0.899 | 0.765 | 0.862 |
| **Qwen3-VL-8B-Instruct-Tuned** | Closed-Set | 0.744 | 0.863 | 0.960 | 0.991 | 0.779 | 0.735 | 0.850 | 0.819 | 0.832 |
| | **Open-Set** | **0.885** | **0.957** | **0.986** | **0.991** | **0.976** | **0.951** | **0.935** | **0.915** | **0.968** |
| InternVL3-8B-Instruct-Tuned | Closed-Set | 0.503 | 0.883 | 0.812 | 0.996 | 0.301 | 0.535 | 0.669 | 0.449 | 0.645 |
| | Open-Set | 0.564 | 0.804 | 0.824 | 0.978 | 0.473 | 0.558 | 0.715 | 0.596 | 0.665 |

Among the adapted models, Qwen3-VL-8B-Instruct-Tuned achieved the best overall performance. Under the closed-set setting, its F1-scores were 0.850 for deer, 0.819 for rhino, and 0.832 for elephant. Under the open-set setting, these values further increased to 0.935, 0.915, and 0.968, respectively, yielding the highest results in Table 3. Qwen2.5-VL-7B-Instruct-Tuned also showed strong and relatively balanced performance, whereas InternVL3-8B-Instruct-Tuned improved substantially over its untuned counterpart but remained weaker overall, especially for rhino and elephant. This difference may be related to the model-specific fine-tuning configuration, as InternVL3-8B-Instruct had the smallest proportion of trainable parameters, whereas Qwen3-VL-8B-Instruct had the largest.

Comparing the two prompting settings, open-set prompting generally produced better recognition results than closed-set prompting for all three tuned models, with the largest improvement observed for Qwen3-VL-8B-Instruct-Tuned. This suggests that open-set prompting is more effective than closed-set prompting for species recognition in the adapted VLM framework.

In summary, these results demonstrate that multimodal projector alignment effectively transfers RGB-pretrained VLMs to drone thermal imagery, and that Qwen3-VL-8B-Instruct-Tuned with open-set prompting provided the most robust species recognition performance in this study.

#### 4.1.2 Instance Enumeration

Table 3 summarizes instance enumeration performance across the foundation and tuned VLMs under closed-set and open-set prompting. Overall, multimodal projector alignment substantially improved counting performance for all three models. Compared with the untuned foundation models, the tuned models achieved higher exact count accuracy and within-1 accuracy, together with lower MAE values across most species.

**Table 3** Comparison of instance enumeration performance across models.

| Model | Prompt | Exact accuracy | | | Within-1 accuracy | | | MAE | | |
|---|---|---|---|---|---|---|---|---|---|---|
| | | Deer | Rhino | Elephant | Deer | Rhino | Elephant | Deer | Rhino | Elephant |
| Qwen2.5-VL-7B-Instruct | Closed-Set | 0.283 | 0.650 | 0.544 | 0.579 | 0.942 | 0.933 | 3.150 | 0.442 | 0.770 |
| | Open-Set | 0.417 | 0.571 | 0.403 | 0.571 | 0.920 | 0.912 | 7.690 | 0.575 | 1.062 |
| Qwen3-VL-8B-Instruct | Closed-Set | 0.376 | 0.739 | 0.726 | 0.633 | 0.951 | 0.876 | 4.199 | 0.327 | 0.491 |
| | Open-Set | 0.247 | 0.526 | 0.345 | 0.606 | 0.871 | 0.849 | 4.323 | 0.726 | 0.849 |
| InternVL3-8B-Instruct | Closed-Set | 0.345 | 0.761 | 0.677 | 0.558 | 0.956 | 0.858 | 4.553 | 0.327 | 0.593 |
| | Open-Set | 0.088 | 0.217 | 0.133 | 0.434 | 0.743 | 0.708 | 5.960 | 1.150 | 1.442 |
| Qwen2.5-VL-7B-Instruct-Tuned | Closed-Set | 0.482 | 0.854 | 0.863 | 0.743 | 0.969 | 0.960 | 1.854 | 0.181 | 0.181 |
| | Open-Set | 0.496 | 0.867 | 0.885 | 0.761 | 0.969 | 0.978 | 1.602 | 0.173 | 0.142 |
| Qwen3-VL-8B-Instruct-Tuned | Closed-Set | 0.580 | 0.889 | 0.938 | 0.770 | 0.987 | 0.987 | 2.265 | 0.137 | 0.075 |
| | **Open-Set** | **0.580** | **0.907** | **0.965** | **0.779** | **0.982** | **1.000** | **2.664** | **0.119** | **0.035** |
| InternVL3-8B-Instruct-Tuned | Closed-Set | 0.513 | 0.863 | 0.796 | 0.735 | 0.978 | 0.951 | 1.743 | 0.168 | 0.318 |
| | Open-Set | 0.504 | 0.841 | 0.708 | 0.739 | 0.987 | 0.894 | 1.956 | 0.173 | 0.593 |

Among the adapted models, Qwen3-VL-8B-Instruct-Tuned achieved the best overall counting performance. Under the open-set setting, it performed particularly well for rhino and elephant, with exact accuracies of 0.907 and 0.965, within-1 accuracies of 0.982 and 1.000, and MAE values of 0.119 and 0.035, respectively. For deer, the exact accuracy remained 0.580, while the within-1 accuracy increased slightly to 0.779. Qwen2.5-VL-7B-Instruct-Tuned also showed strong and stable performance, especially under open-set prompting, with exact accuracies of 0.496 for deer, 0.867 for rhino, and 0.885 for elephant. The corresponding within-1 accuracies were 0.761, 0.969, and 0.978, with relatively low MAE values across all species. InternVL3-8B-Instruct-Tuned improved substantially over its untuned counterpart, but its performance remained less consistent, particularly for elephant under open-set prompting.

Comparing prompting strategies, open-set prompting generally produced better enumeration performance for the tuned Qwen-based models, particularly for rhino and elephant. Overall, these results indicate that multimodal projector alignment effectively improves instance enumeration in drone thermal imagery, and that Qwen3-VL-8B-Instruct-Tuned with open-set prompting provided the most robust counting performance in this study.

### 4.1.3 Qualitative Comparison

Fig. 5 presents representative qualitative results of the fine-tuned models for species recognition and instance enumeration in drone thermal imagery. Across the three examples, Qwen3-VL-8B-Instruct under open-set inference correctly predicted both species and count, whereas the alternative model settings showed species misclassification and/or counting errors. These examples provide qualitative support for the quantitative benchmarking results reported in Tables 2 and 3.

In the deer example, all model settings correctly identified the target species, but substantial differences were observed in enumeration accuracy were observed. Qwen3-VL-8B-Instruct under open-set inference produced the correct count of 10, whereas the remaining models underestimated the number of individuals, with predictions ranging from 1 to 6. In the elephant example, only Qwen3-VL-8B-Instruct under open-set inference correctly predicted *elephant, 1*. The other models confused the target with rhino or deer, although the count of one individual was preserved. In the rhino example, Qwen3-VL-8B-Instruct under open-set inference correctly predicted both species and counts, whereas the remaining models misclassified the animals as deer or elephants, and some also underestimated the counts.

Overall, the result shows that Qwen3-VL-8B-Instruct under open-set inference delivered the most reliable qualitative performance among the evaluated fine-tuned models, with accurate outputs for both species recognition and instance enumeration across all representative cases. These examples further indicate that open-set inference can enhance the practical effectiveness of fine-tuned VLMs for interpreting drone thermal imagery.

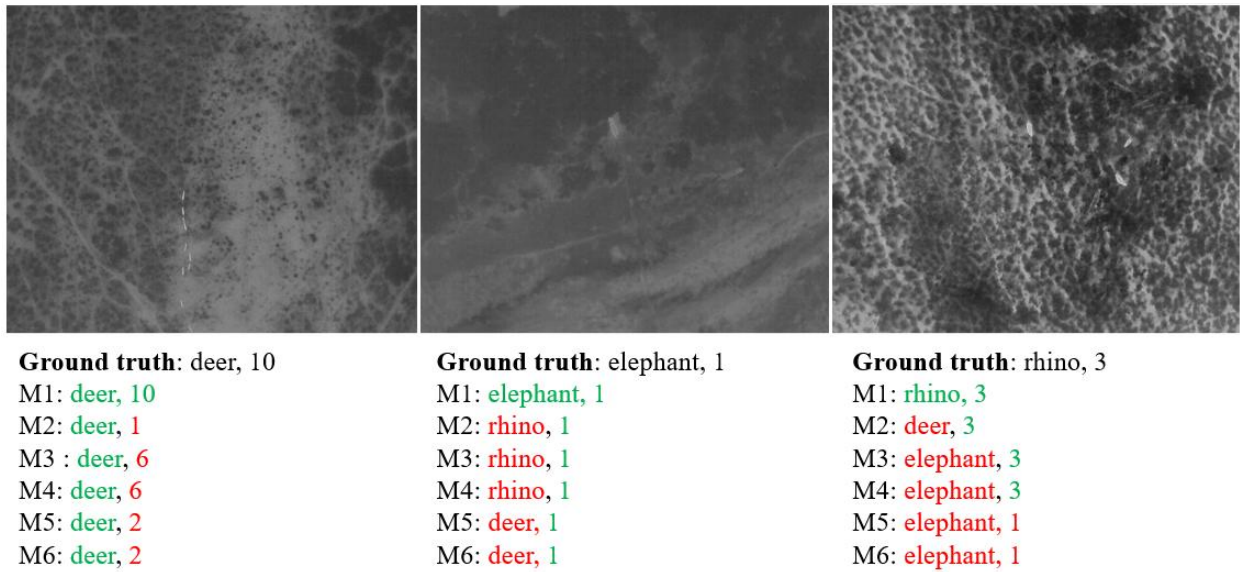

**Figure 5**. Representative results of the fine-tuned models, where **Qwen3-VL-8B-Instruct under open-set inference** correctly achieves species identification and counting, while alternative models produce species and/or count errors. Model abbreviations: **M1**: Qwen3-VL-8B-Instruct (open-set inference); **M2**: Qwen3-VL-8B-Instruct (closed-set inference); **M3**: Qwen2.5-VL-7B-Instruct (open-set inference); **M4**: Qwen2.5-VL-7B-Instruct (closed-set inference); **M5**: InternVL3-8B-Instruct (open-set inference); **M6**: InternVL3-8B-Instruct (closed-set inference).

### 4.2 Best model analysis

Based on the model benchmarking results, Qwen3-VL-8B-instrict achieved the best overall performance in species recognition and instance enumeration with drone thermal imagery. To further examine the optimization behavior of the best performing model, training was conducted for 1,500 steps, and other hyperparameters were kept the same as in benchmarking. Fig. 6 shows the training and validation loss curves during multimodal projector alignment. The blue curves represent the original loss values, whereas the orange curves represent the exponentially smoothed trends.

As shown in Fig. 6(a), the training loss decreased sharply during the early stage of optimization, indicating that the projector rapidly learned to align thermal visual features with the semantic space of the pretrained VLM. After this initial drop, the training loss continued to decline gradually and stabilized at a low level, suggesting effective convergence of the alignment process. In contrast, the validation loss in Fig. 6(b) decreased substantially during the early and middle stages of training, reaching its lowest level at approximately **1000 steps**. After that point, the validation loss gradually increased, while the training loss continued to decline. This divergence suggests the onset of overfitting during the later stage of training, particularly from 1000 to 1500 steps. The results suggest that the checkpoint around **1000 steps** provides the best generalization performance and should therefore be selected as the optimal checkpoint for inference. Overall, the loss curves indicate that multimodal projector alignment provided stable and efficient optimization for Qwen3-VL-8B-Instruct, while also showing that model selection based on validation loss is necessary.

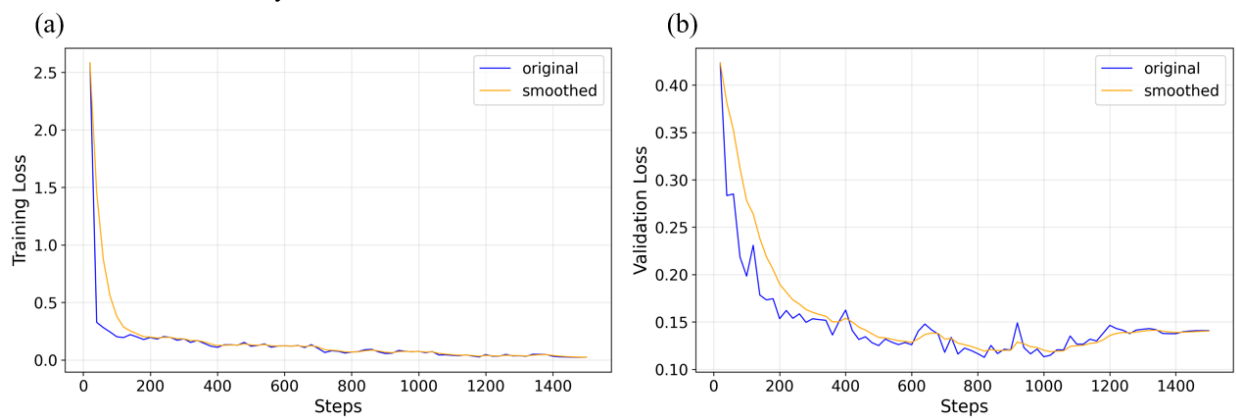

**Figure 6**. (a) Training and (b) validation loss curves of Qwen3-VL-8B-Instruct during fine-tuning.

Fig. 7 compares the inference performance of Qwen3-VL-8B-Instruct using checkpoints at 500, 1000, and 1500 training steps. For species recognition, the 1500-step checkpoint achieved the highest precision and F1-scores overall, but its recall for deer and rhino was slightly lower than that of the 1000-step checkpoint, suggesting that later training improved prediction confidence while reducing sensitivity for some classes. By contrast, the 1000-step checkpoint produced the best recall for deer and rhino, indicating stronger generalization at this stage. For instance, enumeration, the 1500-step checkpoint achieved the lowest MAE, particularly for deer, indicating improved count error reduction. However, it did not consistently yield the best exact counting performance. The 1000-step checkpoint achieved the highest exact accuracy for rhino and the best within-1 accuracy for deer and elephant, demonstrating a more balanced counting performance across species.

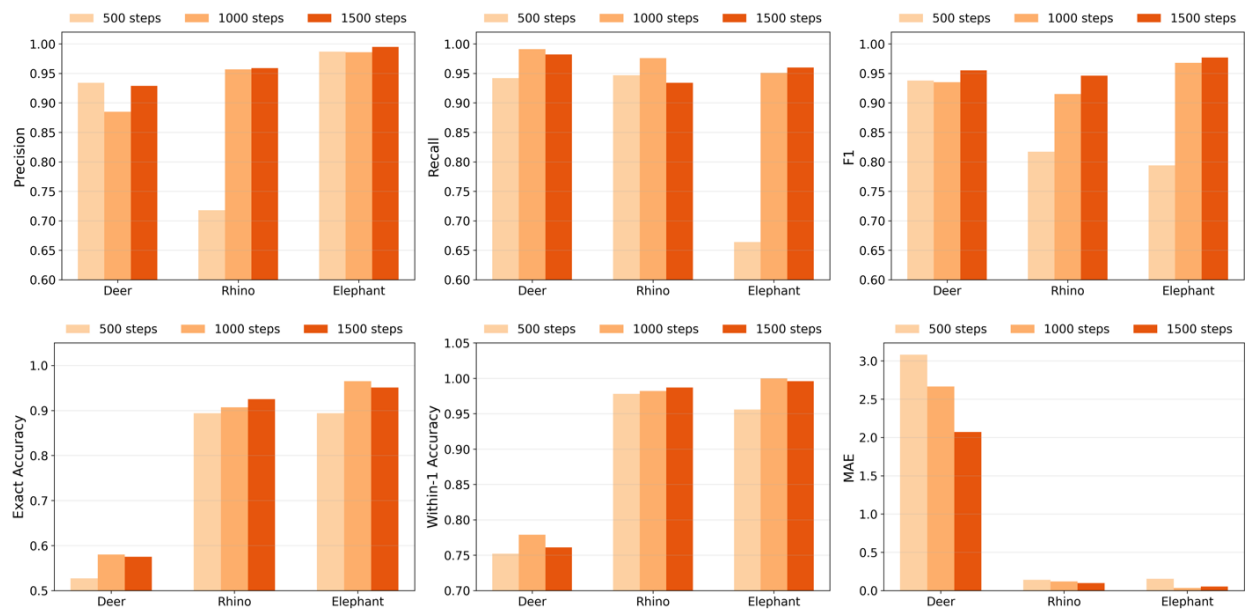

**Figure 7**. Performance of Qwen3-VL-8B-Instruct fine-tuned under varying training step budgets (500, 1000, 1500 iterations).

Overall, although the 1500-step checkpoint provided slightly better precision, F1-score, and MAE in some cases, the 1000-step checkpoint showed stronger validation performance, better recall for key species, and more balanced instance enumeration results. Combined with the validation loss analysis in Fig. 6, these results suggest that the 1000-step checkpoint of Qwen3-VL-8B-Instruct model offers the best generalization performance and should therefore be selected as the optimal model for inference on the thermal drone imagery dataset.

### 4.3 Habitat-Context Interpretation

This section presents the image interpretation performance of the best-performing model, Qwen3-VL-8B-Instruct, using drone-collected thermal and RGB imagery simultaneously. In addition to species recognition and instance enumeration, the model was used to generate habitat-context outputs, including land-cover characteristics, key landscape features, visible human disturbance, and a brief habitat-context interpretation. For RGB imagery, the prompt was defined as follows:

*Describe the most important environmental context in this drone image. Return 4 lines only:*
*Habitat/land cover:*
*Key landscape features (e.g., river, road, forest edge, grassland).*
*Human presence/disturbance (if any).*
*Brief habitat-context interpretation (1 sentence).*

Fig. 8 presents representative outputs generated from paired drone thermal and RGB imagery. In Fig. 8(a), the model correctly identified rhino and counted three individuals from the thermal image, while the RGB image supported habitat-context interpretation of a dense forest environment. The generated output described the scene as dense tropical rainforest with mixed canopy layers, highlighted thick vegetation and tree trunks as the dominant landscape features, and indicated no visible roads, rivers, or direct human infrastructure. In Fig. 8(b), the model correctly identified elephant and counted one individual, while also recognizing the riverine setting from the paired RGB image. The model described the land cover as dense tropical rainforest with a riverine edge and correctly identified the riverbank, dense vegetation, and undisturbed natural setting as the main contextual features.

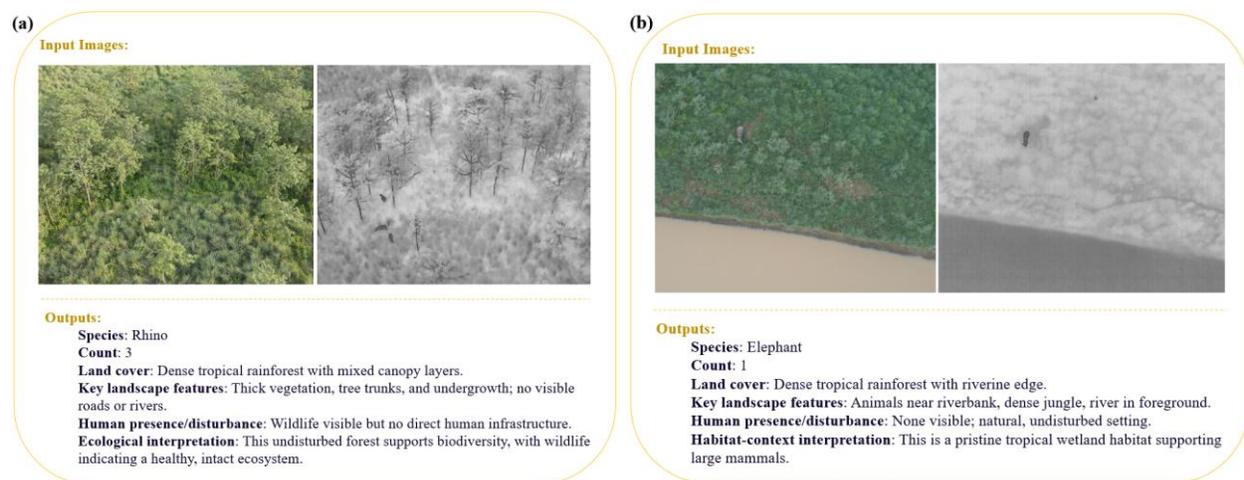

**Figure 8**. Representative outputs of Qwen3-VL-8B-Instruct using paired RGB and thermal imagery in (a) a forested setting and (b) a riverine setting.

Overall, these examples demonstrate that the adapted VLM can integrate thermal and RGB information to generate not only accurate species recognition and instance enumeration, but also habitat-context interpretation by identifying land-cover conditions, key landscape features, and visible human disturbance. They further illustrate the model's practical application for context-aware wildlife monitoring across different environmental settings.

### 6. Conclusion

This study demonstrates that RGB-pretrained vision–language models can be effectively extended to drone thermal imagery through lightweight multimodal adaptation. Among the adapted models, Qwen3-VL-8B-Instruct achieved the best overall performance for species recognition and instance enumeration. Further analysis showed that open-set prompting generally outperformed closed-set prompting and that training duration also influenced inference performance. In particular, the checkpoint at 1000 training steps provided the best balance between efficiency and generalization performance. By integrating simultaneously acquired drone RGB imagery, the model also generated habitat-context interpretation by identifying land-cover conditions, key landscape features, and visible human disturbance.

These findings indicate that the adapted Qwen3-VL-8B-Instruct model can support not only species recognition and instance enumeration from thermal imagery, but also habitat-context interpretation when paired with RGB imagery. This proposed framework thus extends drone-based wildlife analysis beyond target-level prediction toward multimodal, context-aware interpretation of ecological scenes. It therefore provides a useful foundation for future research in context-aware wildlife monitoring, habitat mapping, and broader multimodal ecological interpretation from drone imagery. Future work will extend this framework by incorporating animal behavior analysis to further enrich the ecological information derived from drone imagery.

**Author Contributions**

**Hao Chen**: Writing – original draft, Writing – review & editing, Data curation, Validation, Software, Methodology.
**Fang Qiu**: Writing – review & editing, Supervision, Project administration, Funding acquisition, Conceptualization.
**Fangchao Dong**: Data curation, Visualization. **Defei Yang**: Data curation, Validation, Writing – review & editing.
**Eve Bohnett**: Data curation, Formal analysis. **Li An**: Conceptualization, Funding acquisition.

**Acknowledgements**

We thank the Chitwan National Park, notably the protected area managers and staff who supported the field work.

**Funding Information**

This research was performed with financial support from the National Science Foundation (NSF) of the USA grant number BCS-1826839.

**Conflict of Interest Statement**

The authors declare that they have no known competing financial interests or personal relationships that could have appeared to influence the work reported in this paper.

**Data Availability**

Data will be made available on request.